\pdfoutput=1
\documentclass[11pt]{article}

\usepackage{ACL2023}

\usepackage{times}
\usepackage{latexsym}

\usepackage[T1]{fontenc}
\usepackage[utf8]{inputenc}
\usepackage{hyperref}
\usepackage{multirow}
\usepackage{amsmath}
\usepackage{amssymb}
\usepackage{pifont}
\usepackage{tabularx}
\usepackage{color,colortbl}
\usepackage{graphicx}
\usepackage{placeins}
	
\definecolor{Gray}{gray}{0.9}
\DeclareMathOperator*{\argmax}{arg\,max}  % in your preamble

\usepackage{microtype}

\usepackage{inconsolata}

\title{Towards Weakly-Supervised Hate Speech \\ Classification Across Datasets}

\author{Yiping Jin$^{1}$, Leo Wanner$^{2,1}$, Vishakha Laxman Kadam$^3$, Alexander Shvets$^1$ \\
$^1$NLP Group, Pompeu Fabra University, Barcelona, Spain\\
$^2$Catalan Institute for Research and Advanced Studies\\
$^3$Knorex, 02-129 WeWork Futura, Pune, India\\
\texttt{\{yiping.jin, leo.wanner, alexander.shvets\}@upf.edu} \\
\texttt{vishakha.kadam@knorex.com} \\
}

\begin{document}
\maketitle
\begin{abstract}
As pointed out by several scholars, current research on hate speech (HS) recognition is characterized by unsystematic data creation strategies and diverging annotation schemata. Subsequently, supervised-learning models tend to generalize poorly to datasets they were not trained on, and the performance of the models trained on datasets labeled using different HS taxonomies cannot be compared. To ease this problem, we propose applying extremely weak supervision that only relies on the class name rather than on class samples from the annotated data. We demonstrate the effectiveness of a state-of-the-art weakly-supervised text classification model in various in-dataset and cross-dataset settings. Furthermore, we conduct an in-depth quantitative and qualitative analysis of the source of poor generalizability of HS classification models.

\textbf{Content Warning:} \textit{This document discusses examples of harmful content (hate, abuse, and negative stereotypes). The authors do not support the use of harmful language.}

\end{abstract}

\section{Introduction}
\label{sec:intro}

Due to a growing concern about its impact on society, hate speech  (HS) recognition recently received much attention from the NLP research community~\citep{bilewicz2020hate}. A large number of proposals on how to address HS as a supervised classification task have been put forward; see, among others,~\citep{waseem-hovy-2016-hateful,waseem-2016-racist,poletto2021resources} and several shared tasks have been organized~\citep{basile-etal-2019-semeval,caselli-etal-2020-feel}. 

However, while Transformers models such as BERT~\citep{devlin-etal-2019-bert} achieved impressive performance on various benchmark datasets~\citep{swamy-etal-2019-studying}, recent work demonstrated that state-of-the-art HS classification models generalize poorly to datasets other than the ones they are trained on~\citep{fortuna-etal-2020-toxic,fortuna2021well,yin2021towards}, even when the datasets come from the same data source, e.g., Twitter. This casts a doubt on what we have achieved in the HS classification task.

\citet{fortuna-etal-2022-directions} identify three main challenges related to HS classification: 1. {\it the definitorial challenge}: while the interpretation of what is HS highly depends on the cultural and social norms of its creator \citep{talat_you_2022}, HS research favours a universal definition; 2. {\it the annotation challenge}: due to the subjective nature of HS, the annotation also often depends on  the context, the social bias of the annotator, and their familiarity with the topic~\citep{wiegand-etal-2019-detection}, such that the annotators with different backgrounds tend to provide deviating annotations~\citep{waseem-2016-racist,olteanu2018effect}, especially when not only the presence of HS is to be annotated, but also its category and the group it targets~\citep{basile-etal-2019-semeval}; 3. {\it the learning and evaluation challenge}: the common evaluation practice of the HS classification models assumes that the distributions of the training data and the data to which the model is applied are identical, which is not the case in reality; real-world HS data is relatively rare, while the strategies applied for the creation of HS datasets favor explicit HS expressions~\citep{sap-etal-2020-social,yin2021towards}, using search with explicit target keywords~\cite{waseem-hovy-2016-hateful,basile-etal-2019-semeval}.

%discuss two main challenges to HS classification: \textit{task definition} and \textit{annotation}. 
%While shared tasks and datasets help advance the field, 

%a universal taxonomy for hate speech may not be ideal because the discretion of HS depends on the cultural context and the target group. To this end, they advocate narrowing down the definition of HS instead of making universalist assumptions. Meanwhile, previous work in HS classification also often utilized different taxonomies and focused on different domains. 

%Due to task definition and annotation challenges, we are unlikely to build a one-size-fits-all HS classification model. Instead, an approach that can adapt to different taxonomies and learn from the distribution in unlabeled datasets is highly desirable. 

In order to address these challenges, we propose the use of extremely weak supervision, which uses category names as the only supervision signal~\citep{meng-etal-2020-text,wang-etal-2021-x}: Extremely weak supervision does not presuppose any definition of HS, which would guide the annotation, such that when the interpretation of what is to be considered as HS is modified, we can retrain the model on the same dataset, without the need of re-annotation. Furthermore, when the data distribution changes, the model can learn from unlabeled data and adapt to a new domain. 

Our contributions can be summarized as follows:

\vspace{-\topsep}
\begin{itemize}
  \setlength\itemsep{-0.3em}
  \item We apply extremely weak supervision to HS classification and achieve promising performance compared to fully-supervised and weakly-supervised baselines.
  \item We perform cross-dataset classification under different settings and yield insights on the transferability of HS datasets and models.
  \item We conduct an in-depth analysis and highlight the potentials and limitations of weak supervision for HS classification.
\end{itemize}
\vspace{-\topsep}

\section{Related Work}
\label{sec:literature}

Since our goal is to advance the research on HS classification, we focus, in what follows, on the review of related work in this area and refrain from the discussion of the application of weakly supervised supervision models to other problems.  

Standardizing different hate speech (HS) taxonomies across datasets is a first step in performing cross-dataset analysis and experiments. To this end, \citet{fortuna-etal-2020-toxic} created a category mapping among six publicly available HS datasets. Furthermore, they measured the data similarity of categories in an intra- and inter-dataset manner and reported the performance of a public HS classification API on different datasets and categories.

Other previous work in cross-dataset HS classification followed similar experimental settings by training a supervised classifier on the training set of each dataset and reporting the performance on the corresponding test set and test sets from other datasets. For instance,
\citet{karan-snajder-2018-cross} trained linear SVM models on 9 different HS datasets. They showed that models performed considerably worse on out-of-domain datasets. They further performed domain adaptation using the FEDA framework~\citep{daume-iii-2007-frustratingly} and demonstrated that having at least some in-domain data is crucial for achieving good performance. Similarly, \citet{swamy-etal-2019-studying} compared Linear SVM, LSTM, and BERT models trained on different datasets. They reported that some pairs of datasets perform well on each other, likely due to a high degree of overlap. They also claimed that a more balanced class ratio is essential for the datasets' generalizability.

\citet{fortuna2021well} conducted a large-scale cross-dataset experiment by training a total of 1,698 classifiers using different algorithms, datasets, and other experimental setups. They demonstrated that the generalizability does not only depend on the dataset but also on the model. Transformer-based models have a better potential to generalize to other datasets, likely thanks to the wealth of data they have observed during pre-training. Furthermore, they built a random forest classifier to predict the generalizability based on human-engineered dataset features. The experiment revealed that to achieve cross-dataset generalization, the model must first perform well in an intra-dataset scenario. In addition, inconsistency in class definition hampers generalizability.

\citet{wiegand-etal-2019-detection} and \citet{arango2019hate} studied the impact of data bias on the generalizability of HS models, with the outcome that  popular benchmark datasets possess several sources of biases, such as bias towards explicit HS expressions, topic bias, and author bias. The classification results dropped significantly when the bias is reduced. To this end, they proposed using cross-dataset classification as a way to evaluate models' performance in a more realistic setting.

\citet{gao-etal-2017-recognizing} argued that the low frequency of online HS impedes obtaining a wide-coverage HS detection dataset. To this end, they proposed a two-path bootstrapping approach involving an explicit slur term learner and an LSTM~\citep{hochreiter1997long} classifier. The slur term learner is initialized with a list of hand-engineered seed slur terms and applies to an unlabeled dataset to automatically label hateful posts, which are used to train the classifier. The slur term learner and the classifier are trained iteratively in a co-training manner~\citep{blum1998combining}. 

A distinct approach was proposed by \citet{talat2018bridging}. This approach utilized multi-task learning (MTL) to enhance domain robustness. They trained a classifier on three distinct sets of annotations: \citet{waseem-hovy-2016-hateful}, \citet{waseem-2016-racist}, and \citet{davidson2017automated}. While MTL helps to prevent overfitting and may provide auxiliary fine-grained predictions, it requires annotating a dataset using different taxonomies, granularities, or aspects.

%In a recent position paper, \citet{fortuna-etal-2022-directions} highlights that the supervised classification framework is poorly suited for HS due to its complex and subjective nature. Specifically, Supervised classification requires coining a clear \textit{task definition}, obtaining unbiased and high-quality \textit{annotation}, and objective \textit{evaluation} of models' performance and generalizability. Each of these steps remain an open research problem. Our work can be regarded as an answer to \citet{fortuna-etal-2022-directions}'s call for alternative conceptualization for HS identification.

Our approach is most similar to \citet{jin2022learning}'s, which also applied weakly-supervised learning on a target-domain dataset. However, their approach requires mining a list of 30 high-quality keywords for each category from a large labeled source-domain dataset. Moreover, they assume that the source and target datasets are labeled using the same HS taxonomy.

\section{Weakly-Supervised HS Classification}
\label{sec:method}

In this section, we briefly introduce the basics of weakly supervised text classification and then discuss  the cross-dataset classification we aim for.

\subsection{Preliminaries: Weakly Supervised Text Classification}
\label{subsec:preliminaries}

Weakly-supervised text classification eliminates the need for a large labeled dataset~\citep{meng2018weakly,mekala-shang-2020-contextualized}. Instead, it trains classifiers using a handful of labeled seed words and unlabeled documents. While the human annotation effort is significantly reduced, weakly-supervised classification methods are sensitive to the choice of seed words, and the process to nominate high-quality seed words is non-trivial~\citep{jin-etal-2021-seed}.

More recently, \citet{meng-etal-2020-text} and \citet{wang-etal-2021-x} explored \textit{extremely} weak supervision, where the model is given only the category name instead of manually curated seed words. Extremely weak supervision is well suited for hate speech detection because we may not know all the aspects of hate speech for a particular category or target group, or what a user may interpret as a HS statement that falls into a specific category. On top of that, extremely weak supervision often performs semantic expansion on the unlabeled dataset and automatically augments the category representation with new aspects (in the form of seed words).

We choose X-Class~\citep{wang-etal-2021-x} as the primary weakly-supervised classification method, which matches or outperforms previous state-of-the-art weakly-supervised methods on 7 benchmark datasets. 
X-Class first estimates category representations by incorporating words similar to each category. It represents each word by its averaged contextualized word embedding across the entire dataset. It then iteratively adds the most similar word to each category whose embedding has the highest cosine similarity to the category representation. The category representation is updated as a weighted average of the expanded keywords. Expressly, they assume a Zipf's law distribution~\citep{powers1998applications} and weight the $j$-th keyword by 1/$j$.

\begin{equation} \label{eq:class-repre}
s_\ell = \frac{\sum^{|\kappa_\ell|}_{j=1}1/j \cdot s_{\kappa_{\ell,j}}}{\sum^{|\kappa_\ell|}_{j=1}1/j}
\end{equation}

\noindent where $\boldsymbol{\kappa_{\ell,j}}$ is the $j$-th keyword of category $\boldsymbol{\ell}$ and $\boldsymbol{s_{\kappa_{\ell,j}}}$ is its average contextualized embedding. X\nobreakdash-Class also performs a consistency check and stops adding new words if a category's nearest words have changed.

Then, X-Class derives document $\boldsymbol{i}$'s category-oriented document representation $\boldsymbol{d_i}$ by weighting each word in the document based on its similarity to the category representations. Afterward, it clusters the documents using a Gaussian Mixture Model (GMM)~\citep{duda1973pattern} by initializing the category representations as cluster centroids. Finally, the most confident pseudo-labeled documents from each cluster are used to train a text classifier.

In our initial experiments, we observed that while GMM generally improves the pseudo-labeling, the accuracy for some low-frequency categories tends to drop sharply. This is likely because GMM works as a \textit{global} density estimator. Therefore, data of the more frequent categories may ``attract'' more weights and cause the category representation for low-frequency categories to diverge too much from its initial category representation. To address this problem, we introduce an additional \textit{representation}-based prediction, which assigns document $\boldsymbol{i}$ to the category representation which has the highest cosine similarity:

\begin{equation} \label{eq:repr_pred}
\ell_{i}^{rep} = \argmax_{\ell \in L}cosine(s_\ell, d_i)
\end{equation}

\noindent We denote GMM's category assignment for document $\boldsymbol{i}$ as `$\ell_{i}^{gmm}$'. Instead of pseudo-labeling most confident documents based on GMM only, we take the subset of confident documents to which GMM and representation-based prediction assign the same label ($\ell_{i}^{gmm} = \ell_{i}^{rep}$). This ensures that the document is sufficiently close to the original category representation. We denote this modified version as `X-Class$^{Agree}$'.

\subsection{Cross-Dataset Classification}
\label{subsec:out-domain}

In this work, we study cross-dataset classification, where we do not have any document labels in the target dataset. A dataset is characterized by its \textit{documents} (and their underlying topics and word distributions) and \textit{taxonomy} (list of categories).\footnote{While the term ``cross-domain'' is more popular than ``cross-dataset'', it does not suggest that the source and target dataset's taxonomies may differ. The discussion of the related problem of cross-task generalization~\citep{raffel2020exploring,sanh2022multitask}, which works for unrelated tasks, is beyond the scope of this work.} 

Given a single HS dataset with its corresponding categories, we can straightforwardly apply X-Class using the category names and an unlabeled dataset. On the other hand, both the data distribution and taxonomy may differ when we experiment on different datasets.
%Due to data sparsity, researchers usually apply sampling methods to create hate speech datasets, such as searching by keywords, specific hashtags, or user accounts. As a result, supervised HS detection models generalize poorly even to datasets collected from the same data source~\cite{swamy-etal-2019-studying,fortuna2021well,yin2021towards}. 
There are three different cases for the relation between the taxonomies of the source and target datasets.

\vspace{-\topsep}
\begin{itemize}
  \setlength\itemsep{-0.3em}
  \item \textbf{1-to-1:} The target taxonomy is identical to the source taxonomy or a subset of it. 
  \item \textbf{N-to-1:} The target taxonomy differs from the source taxonomy, but each target category can be mapped to one or more source categories.
  \item \textbf{N-to-N:} The target taxonomy differs from the source taxonomy, and some target categories cannot be mapped to any source category.
\end{itemize}
\vspace{-\topsep}

Supervised learning can be applied in the first two cases: We can create a category mapping from the target categories to the source categories, then use this mapping to either \textit{post-process} the model predictions (converting predicted source categories to target categories) or relabel the dataset using the target taxonomy and \textit{retrain} the model. However, in the last case, we cannot directly apply supervised learning without further data collection and annotation because we lack labeled data for at least some categories. In contrast, weakly-supervised methods do not require labeled documents and can readily utilize unlabeled documents in the target dataset to capture the underlying distribution. Furthermore, even when applied to a completely unseen dataset, it can also ``relabel'' the source dataset using the target taxonomy and bootstrap a classifier.

\section{Experiments}
\label{sec:exp}

\subsection{Datasets}
\label{subsec:datasets}

We conduct experiments on two popular HS datasets that differ with respect to the data source and taxonomy of HS categories: the Waseem dataset and the SBIC dataset. The Waseem dataset~\citep{waseem-hovy-2016-hateful}\footnote{\url{https://github.com/zeeraktalat/hatespeech}} contains 5,355 tweets with sexist and racist content. The dataset was annotated by the authors (inter-annotator agreement $\kappa=0.84$) and reviewed by a domain expert (a gender studies student who is a non-activist feminist). The SBIC dataset~\citep{sap-etal-2020-social}\footnote{\url{https://maartensap.com/social-bias-frames/}} contains 44,671 posts collected from different domains: Reddit, Twitter, and hate sites. It was annotated by crowdsource workers on Amazon Mechanical Turk. A small portion of the data is originally from the Waseem dataset (1,816 posts). We exclude these posts to avoid overlap between the two datasets. 

SBIC dataset does not set a predefined taxonomy for HS categories. Instead, annotators can indicate the target group with free-text answers. We select the most frequent six target groups that can be mapped to the categories in the Waseem dataset. While our proposed weakly-supervised learning method does not depend on category mapping, we select the SBIC categories that can be mapped to compare with supervised learning baselines. Table~\ref{tab:category-mapping} shows this category mapping. 

\begin{table}[!htbp]
\centering
\begin{tabularx}{\textwidth}{p{1.5cm}p{5.0cm}}
\cline{1-2}
\textbf{Waseem} & \textbf{SBIC} \\ \cline{1-2}
Sexist & Women; LGBT \\
\rowcolor{Gray}
Racist & Black; Jewish; Muslim; Asian \\\cline{1-2}
\cline{1-2}
\end{tabularx}
\caption{Category mapping between the Waseem and SBIC datasets.}
\label{tab:category-mapping}
\end{table} 

We use the original train/dev/test split (75\%/12.5\%/12.5\%) in the SBIC dataset and randomly split the Waseem dataset to 90\%/10\% into training and test sets. We apply standard preprocessing following \citet{barbieri-etal-2020-tweeteval}, including user mention anonymization and website links and emoji removal. Table~\ref{tab:dataset-stats} presents the distribution of the posts in the two datasets.

\begin{table}[!htbp]
\centering
\begin{tabularx}{\textwidth}{p{1.6cm}p{2.3cm}rr}
\cline{1-4}
\textbf{Dataset} & \textbf{Category} & \textbf{\# Train} & \textbf{\# Test} \\ \cline{1-4}
\multirow{2}{*}{Waseem} & Sexist & 3,107 & 323  \\
& Racist & 1,799 & 177 \\ \cline{2-4}
&\textit{\textbf{Subtotal}} & 4,906 & 500 \\ \cline{1-4}
\multirow{6}{*}{SBIC} & Women & 2,594 & 351  \\
& Black folks & 2,512 & 576\\ 
& Jewish folks & 847 & 207 \\ 
& LGBT folks & 490 & 53 \\
& Muslim folks & 412 & 85 \\ 
& Asian folks & 224 & 34 \\  
\cline{2-4}
&\textit{\textbf{Subtotal}} & 7,079 & 1,306 \\ \cline{1-4}
\end{tabularx}
\caption{Distribution of the documents per dataset, with the posts that contain no word after post-processing removed. The average number of words per post in the Waseem dataset is 17.1 and in the SBIC dataset 20.0.}
\label{tab:dataset-stats}
\end{table}

\subsection{Compared Methods}
\label{subsec:baselines}

We compare X-Class with two representative supervised learning baselines which are trained using the full \textit{labeled} training dataset:

\vspace{-\topsep}
\begin{itemize}
  \setlength\itemsep{-0.3em}
  \item \textbf{Support Vector Machines (SVM)}~\citep{cortes1995support}\textbf{:} We use scikit-learn's\footnote{\url{https://scikit-learn.org}} linear SGD classifier with default hyper-parameters and tf-idf weighting. 
  \item \textbf{BERT}~\citep{devlin-etal-2019-bert}\textbf{:} We fine-tune the \texttt{bert-base-uncased} checkpoint\footnote{\url{https://huggingface.co/bert-base-uncased}} using the exact hyper-parameters to train the final classifier in X-Class (detailed in Section~\ref{subsec:settings}).
\end{itemize}
\vspace{-\topsep}

We also compare the performance of our model with the following baselines that do not require any document labeling:\footnote{We provide the weakly-supervised learning baselines the full \textit{unlabeled} training dataset for keyword expansion and pseudo-labeling.}

\vspace{-\topsep}
\begin{itemize}
  \setlength\itemsep{-0.3em}
  \item \textbf{Majority class:} Always predict the most frequent category in the training dataset.
  \item \textbf{Keyword voting (category name):} Assign the category whose category name occurs most frequently in the document. Fall back to the majority class prediction if there is a tie or none of the keywords appear.
  \item \textbf{Keyword voting (X-Class keywords):} Same as above, but use the expanded keywords in X\nobreakdash-Class's category representation and their associated weights. Assign the category which receives the highest score.
  \item \textbf{Zero-shot PET}~\citep{schick-schutze-2021-exploiting}\textbf{:} Prompting a pre-trained BERT model using hand-crafted patterns and verbalizers to classify documents. We provide details of this baseline in Appendix~\ref{appendix:reproducibility}.
  \item \textbf{\textsc{W{\scriptsize E}STC{\scriptsize LASS}}}~\citep{meng2018weakly}\footnote{\url{https://github.com/yumeng5/WeSTClass}}\textbf{:} CNN-based neural text classifier. It first generates pseudo documents with a generative model seeded with user-provided keywords for pre-training, then conducts self-training to bootstrap from unlabeled documents. We use three manually curated seed words for each category following \citet{meng2018weakly}.
  \item \textbf{LOTClass}~\citep{meng-etal-2020-text}\footnote{\url{https://github.com/yumeng5/LOTClass}}\textbf{:} A strong baseline using extremely weak supervision. The model first uses a masked language model to expand keywords from the category names, then mines category-indicative words using a novel masked category prediction task. Finally, it generalizes via self-training.
\end{itemize}
\vspace{-\topsep}

\subsection{Experiment Settings}
\label{subsec:settings}

We use the official implementation of X\nobreakdash-Class.\footnote{\url{https://github.com/ZihanWangKi/XClass}} The \texttt{bert-base-uncased} checkpoint is used to calculate the document representation and fine-tune the final classifier; the maximum number of keywords for each category is set to 100; and the 50\% most confident pseudo-labeled documents from each category are used to train the final classifier.

To facilitate a fair comparison with supervised learning methods, we reimplemented the final classifier fine-tuning step using the HuggingFace Transformers trainer\footnote{\url{https://huggingface.co/docs/transformers/main/training}} and performed a minimum manual hyper-parameter tuning (\texttt{learning\_rate=2e-5}; \texttt{num\_epochs=6}; \texttt{weight\_decay=0.05}) on the SBIC dev set and applied them on both datasets. We set the \texttt{max\_length} and \texttt{batch\_size} to 64.

We merged the following original target groups in the SBIC corpus into ``LGBT folks'': ``gay men'', ``lesbian women, gay men'', ``lesbian women'', ``trans women, trans men'', ``trans women''. Table~\ref{tab:keywords} presents the category names used by the models. We use the original category name except for ``LGBT'' because it does not occur in the dataset. Instead, we use ``gay'', the most frequently targeted subgroup in the dataset. As shown in Appendix~\ref{appendix:full-keywords}, X\nobreakdash-Class expands to keywords representing other subgroups in the LGBT community. 

\begin{table}[!htbp]
\centering
\begin{tabularx}{\textwidth}{p{1.2cm}p{1.2cm}rp{2.5cm}}
\cline{1-4}
\textbf{Class} & \textbf{Seed} & \textbf{Count} & \textbf{\textsc{W{\scriptsize E}STC{\scriptsize LASS}}} \\ \cline{1-4}
Sexist & sexist & 1,071 & sexist sexism \newline misogynist \\
\rowcolor{Gray}
Racist & racist & 33 & racist racists racism\\ \cline{1-4}
Women & women & 652 & women woman female\\
\rowcolor{Gray}
Black & black & 1,601 & black blacks \newline n*gro \\ 
Jewish & jewish & 142 & jewish jews jew\\ 
\rowcolor{Gray}
LGBT & gay & 209 & gay gays \newline homosexual\\
Muslim & muslim & 228 & muslim muslims \newline islamic\\ 
\rowcolor{Gray}
Asian & asian & 121 & asian asians \newline chinese \\\cline{1-4}
\cline{1-4}
\end{tabularx}
\caption{Seed words used for each category and their frequency in the training dataset. We manually curated the seed words in X-Class's category representation and select the top-3 ranked keywords to train \textsc{W{\scriptsize E}STC{\scriptsize LASS}}.}
\label{tab:keywords}
\end{table} 

\subsection{Results of the Experiments}
\label{subsec:result}

We report the accuracy and macro P/R/F$_1$ scores to quantify each method's performance.

\textbf{In-Dataset Classification.}\; We first validate the efficacy of the methods using the standard in-dataset setting, providing the corresponding training and test datasets. Table~\ref{tab:result-in-domain} displays the result.

\begin{table}[!htbp]
\centering 
\begin{tabular}{p{0.20\textwidth}cc}
% \begin{tabular}
\hline
\rowcolor{Gray}
\multicolumn{3}{c}{\textbf{Waseem Dataset}} \\\cline{1-3} 
\textbf{Model} & \textbf{Acc} & \textbf{P/R/F$_1$}\\ \cline{1-3}
SVM & 97.2 & 97.1/96.8/96.9  \\
BERT & \textbf{98.2} & \textbf{98.2/97.8/98.0} \\
\cline{1-3}
Majority class & 64.6 & 33.2/50.0/39.2 \\ 
KV (class name) & 64.6 & 57.3/50.1/39.8  \\ 
KV (X-Class) & 67.0 & 76.9/53.6/47.0  \\ 
Zero-shot PET & 49.2 & 66.7/59.9/47.3 \\
\textsc{W{\scriptsize E}STC{\scriptsize LASS}} & 77.8 & 77.8/80.4/77.3  \\
LOTClass & 63.2 & 71.3/70.2/63.2  \\\cline{1-3}
X-Class & 96.2 & 96.9/94.9/95.8  \\
X-Class$^{Agree}$ & \textbf{96.6} & \textbf{97.5/95.2/96.2}  \\
\hline
\rowcolor{Gray}
\multicolumn{3}{c}{\textbf{SBIC Dataset}} \\\cline{1-3} 
\textbf{Model} & \textbf{Acc} & \textbf{P/R/F$_1$}\\ \cline{1-3}
SVM & 90.7 & 93.2/82.5/86.7  \\
BERT & \textbf{95.7} & \textbf{94.2/95.1/94.6} \\
\cline{1-3}
Majority class & 26.9 & 4.5/16.7/7.1 \\ 
KV (class name) & 57.7 & \textbf{85.2}/39.7/41.9  \\ 
KV (X-Class) & 55.2 & 47.8/45.1/40.8  \\ 
Zero-shot PET & 35.1 & 38.4/21.6/15.8 \\
\textsc{W{\scriptsize E}STC{\scriptsize LASS}} & 36.4 & 35.9/34.5/29.9  \\
LOTClass & 54.2 & 29.2/29.3/27.5  \\\cline{1-3}
X-Class & 79.8 & 74.0/81.8/74.8 \\
X-Class$^{Agree}$ & \textbf{81.4} & 76.1/\textbf{85.3/76.6} \\
\cline{1-3}
\end{tabular}
\caption{In-Dataset performance of various models. We highlight the best performances of supervised and weakly-supervised methods in bold.}
\label{tab:result-in-domain}
\end{table}

As expected, BERT outperformed SVM among the supervised-learning baselines on both datasets. Interestingly, keyword voting using only the category name achieved high precision for the SBIC dataset. However, its recall is much lower than that of X\nobreakdash-Class due to variations of expressions within the same category. Using X-Class keywords improved keyword voting's recall by 3.5\% and 5.4\% on the two datasets. However, the precision dropped significantly on the SBIC dataset, likely due to the noisier keywords.

\textsc{W{\scriptsize E}STC{\scriptsize LASS}} performs superior to keyword voting baselines on the Waseem dataset, primarily due to its high recall of the ``Racist'' category. This demonstrates the advantage of semantic representation in neural models. However, its performance pales on the SBIC dataset, revealing its weakness in handling more complex cases that involve class imbalance and overlapping, which has been discussed in \citet{wang-etal-2021-x} and \citet{jin2022learning}. LOTClass demonstrates a similar trend, but performs worse on both datasets.\footnote{LOTClass has a higher accuracy on SBIC dataset because it predicts the vast majority of the documents to the most frequent categories ``Women'' and ``Black''. } We analyze the pseudo-labeling accuracy of weakly-supervised baselines and X\nobreakdash-Class in Appendix~\ref{appendix:case-study}.

Comparing X-Class and X-Class$^{Agree}$, we can see that our modification consistently improved the performance.

\noindent\textbf{Cross-Dataset Classification.}\; We conduct cross-dataset classification using the strongest supervised and weakly-supervised models and show the result in Table~\ref{tab:result-cross-domain}. Note that for the ``Waseem $\rightarrow$ SBIC'' setting, we cannot create a category mapping since the target dataset has more fine-grained categories. Therefore, supervised methods and X\nobreakdash-Class using category mapping to post-process the predictions are not applicable. 

When we train BERT and X\nobreakdash-Class using only source-dataset documents, they both perform worse on the target dataset than the in-dataset results in Table~\ref{tab:result-in-domain}. The performance drop is smaller for ``SBIC $\rightarrow$ Waseem'', likely because SBIC dataset contains representative posts for the Waseem categories. 

\begin{table}[!htbp]
\centering 
\begin{tabular}{p{0.22\textwidth}cc}
% \begin{tabular}
\hline
\rowcolor{Gray}
\multicolumn{3}{c}{\textbf{SBIC $\rightarrow$ Waseem}} \\\cline{1-3} 
\textbf{Model} & \textbf{Acc} & \textbf{P/R/F$_1$}\\ \cline{1-3}
BERT \footnotesize{(post-process)} & \textbf{93.6} & \textbf{92.4}/\textbf{94.7}/\textbf{93.2}  \\
BERT \footnotesize{(retrain)} & \textbf{93.6} & \textbf{92.4}/94.2/\textbf{93.2} \\
\cline{1-3}
X-Class \footnotesize{(post-process)} & 91.6 & 93.5/88.5/90.3 \\ 
X-Class \footnotesize{(retrain)} & 84.4 & 89.9/78.1/80.6  \\ 
X-Class$^{Agree}$ \scriptsize{(post-process)} & \textbf{92.8} & \textbf{94.5}/90.1/\textbf{91.8} \\ 
X-Class$^{Agree}$ \footnotesize{(retrain)} & 92.6 & 93.4/\textbf{90.4}/91.6  \\ 
\hline
\rowcolor{Gray}
\multicolumn{3}{c}{\textbf{Waseem $\rightarrow$ SBIC}} \\\cline{1-3} 
\textbf{Model} & \textbf{Acc} & \textbf{P/R/F$_1$}\\ \cline{1-3}
X-Class \footnotesize{(retrain)} & 60.7 & 61.3/59.8/54.5  \\ 
X-Class$^{Agree}$ \footnotesize{(retrain)} & \textbf{69.8} & \textbf{62.7}/\textbf{62.2}/\textbf{58.3}  \\ 
\cline{1-3}
\end{tabular}
\caption{Cross-dataset performance of BERT and X-Class. Both models are trained using source dataset documents and tested on the target dataset. We highlight the best performances of supervised and weakly-supervised methods in bold.}
\label{tab:result-cross-domain}
\end{table}

Surprisingly, \textit{retraining} the models using the target taxonomy does not outperform \textit{post-processing} using category mapping. However, when a category mapping is unavailable (as in the ``Waseem $\rightarrow$ SBIC'' case), retraining a weakly-supervised classifier using the target taxonomy is the only option for cross-dataset classification without manually annotating more data.

An advantage of weakly-supervised methods is that they can utilize \textit{unlabeled} documents from the target dataset when they are available. Although X\nobreakdash-Class$^{Agree}$ still underperforms BERT when both are trained using the source dataset in the ``SBIC $\rightarrow$ Waseem'' experiment, it surpasses BERT by 3\% in both accuracy and macro F$_1$ score when using unlabeled target-dataset documents~\footnote{We can train weakly-supervised models using unlabeled target dataset, which is equivalent to the in-dataset setting (the X-Class$^{Agree}$ row in Table~\ref{tab:result-in-domain}).}.

Again, X-Class$^{Agree}$ outperforms X-Class in all cases. Subsequently, we use X-Class to refer to X-Class$^{Agree}$ for brevity.

\subsection{Analysis: What Makes Cross-Dataset Classification Challenging?}
\label{sec:Analysis}

As shown in Table~\ref{tab:result-cross-domain}, X-Class's performance dropped significantly in the ``Waseem $\rightarrow$ SBIC'' cross-dataset setting compared to using the SBIC training set. In this section, we try to uncover the causes of the performance drop.

\begin{figure}[!ht]
  \centering \includegraphics[width=219pt]{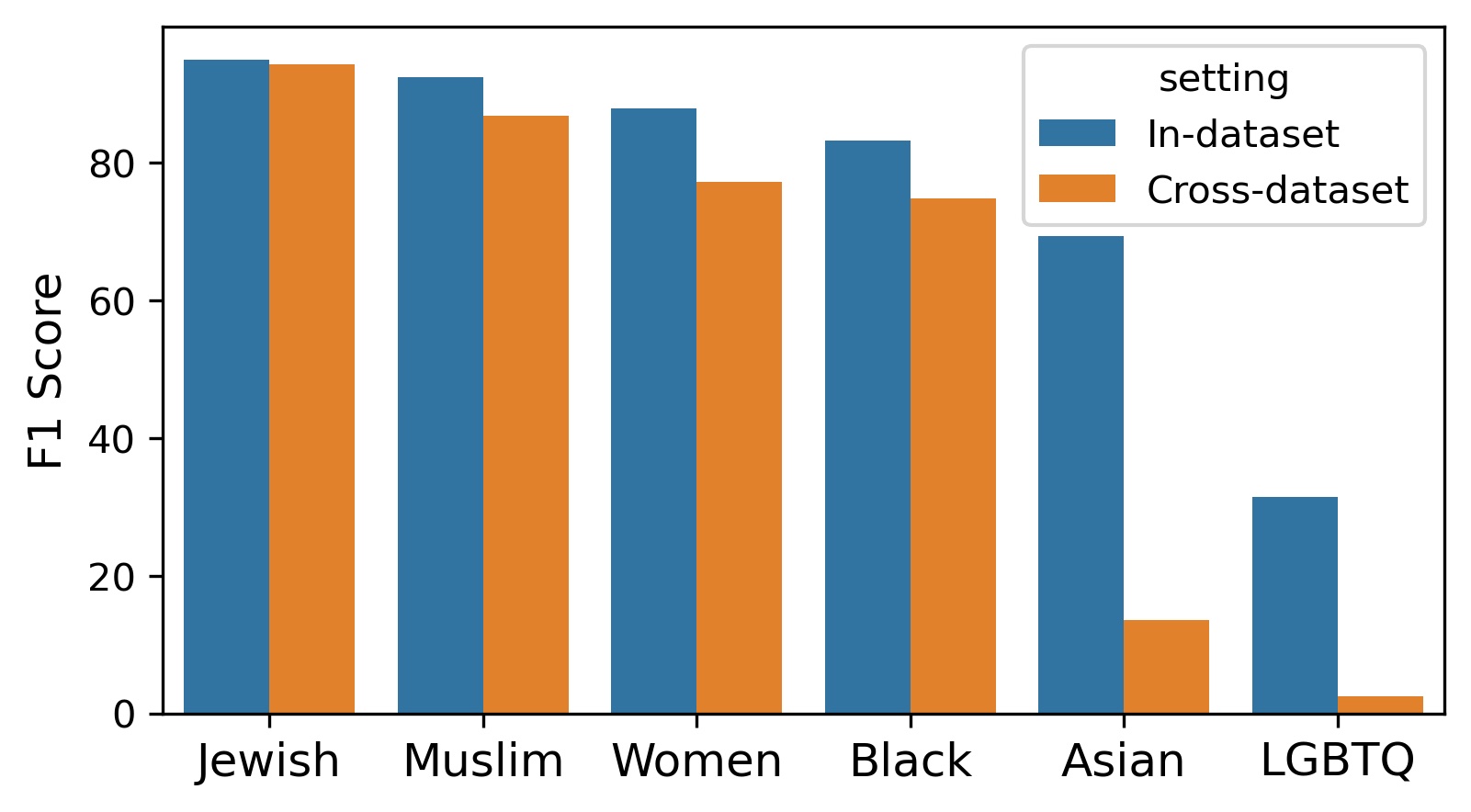}
  \caption{Comparing cross-dataset and in-dataset F$_1$ score of X-Class on the SBIC dataset.}
  \label{fig:category_performance}
\end{figure}

We first plot the per-category F$_1$ score in Figure~\ref{fig:category_performance}. We can see that the cross-dataset model achieved comparable performance as the in-dataset model for the four categories \{Jewish, Muslim, Women, Black\}. However, it failed in the two categories \{Asian, LGBT\}.

\noindent\textbf{Relevant unlabeled documents.}\; Although the Waseem dataset is labeled using a more coarse-grained taxonomy, it may contain documents relevant to some (but not all) fine-grained SBIC categories. Weak supervision usually pseudo-labels the \textit{unlabeled} dataset to train a final classifier. Therefore, it will likely fail when documents related to a particular category are absent in the unlabeled dataset. We count the frequency of documents containing each category name in both datasets and present the results in Table~\ref{tab:keyword-freq}.

\begin{table}[!htbp]
\centering
\begin{tabularx}{\textwidth}{p{1.2cm}p{1.2cm}rr}
\cline{1-4}
\textbf{Class} & \textbf{Seed} & \textbf{Waseem \%} & \textbf{SBIC \%} \\ \cline{1-4}
Sexist & sexist & 21.83\% & 2.70\% \\
Racist & racist & 0.67\% & 1.12\% \\ \cline{1-4}
Women & women & 11.94\% & 9.21\% \\
Black & black & 0.63\% & 22.6\% \\ 
Jewish & jewish & 0.51\% & 2.00\%\\ 
LGBT & gay & 0.59\% & 2.95\% \\
Muslim & muslim & 10.40\% & 3.22\%\\ 
Asian & asian & 0.08\% & 1.71\% \\\cline{1-4}
\cline{1-4}
\end{tabularx}
\caption{Frequency of each category name appearing in the Waseem and SBIC training datasets.}
\label{tab:keyword-freq}
\end{table} 

We can observe that the ``Asian'' category (from the SBIC dataset) is severely under-represented in the Waseem dataset. The word ``Asian'' occurs only 4 times, all in the context of ``Asian women/girls''. 

\citet{waseem-hovy-2016-hateful} conducted a lexical analysis and showed that their ``Sexist'' category is highly skewed towards \textit{women}, and their ``Racist'' category is highly skewed towards \textit{Muslims} and \textit{Jews}.\footnote{Although the term ``Jewish'' has a low frequency, ``Jews'' appears in the ten most frequent terms of the ``Racist'' category.} Coincidentally, these categories also perform the best in the ``Waseem $\rightarrow$ SBIC'' setting.

\begin{table*}[!t]
\centering 
\begin{tabular}{lp{1.0cm}cp{1.0cm}c}
\cline{1-5}
\multirow{2}{*}{Model} &\multicolumn{2}{c}{\textbf{SBIC $\rightarrow$ Waseem}} & \multicolumn{2}{c}{\textbf{Waseem $\rightarrow$ SBIC}}\\ 
&  \textbf{Acc} & \textbf{P/P/F$_1$} & \textbf{Acc} & \textbf{P/R/F$_1$} \\ \cline{1-5}
X-Class (src data \& src category repr) & 92.6 & 93.4/90.4/91.6 & 69.8 & 62.7/62.2/.58.3 \\
X-Class (src data \& tgt category repr) & 93.4 & 92.2/94.0/92.9 & 75.1 & 65.2/55.5/57.8 \\
X-Class (tgt data \& tgt category repr) & 96.6 & 97.5/95.2/96.2 & 81.4 & 76.1/85.3/76.6 \\\cline{1-5}
\end{tabular}
\caption{Cross-dataset performance of X-Class using different unlabeled datasets and category representations.}
\label{tab:cross-domain-deep-dive}
\end{table*}

\noindent\textbf{Category understanding.}\; \citet{jin-etal-2021-seed} argued that weakly-supervised classification and keyword mining are intrinsically related. The failure to identify relevant keywords will harm the category representation and, thus, the classification accuracy. Appendix~\ref{appendix:full-keywords} presents the full list of keywords X\nobreakdash-Class added to the category representations in both in-dataset and cross-dataset settings.

A general observation is that X-Class tends to include fewer keywords in its category representation in the cross-dataset setting. Recall that it stops adding keywords once the consistency check is violated. We hypothesize that the mismatch between the dataset and the taxonomy caused the mined keywords to be noisier and more likely to fail the consistency check.

The four categories that perform the best in both in-dataset and out-dataset settings also contain better-quality keywords. In contrast, the ``Asian'' category's keyword in the cross-dataset setting is entirely off-topic due to its rare occurrence and collocation with words like ``women'' or ``girls''. The ``LGBT'' category contains many vulgar keywords with sexual references, which caused it to confuse with the ``Women'' category.

\noindent\textbf{Class definition vs. dataset.}\; Previous studies tried to explain why HS classification models generalize poorly across datasets, the most frequently cited reasons being the lack of a standardized definition of hate speech~\citep{waseem-hovy-2016-hateful,fortuna-etal-2020-toxic,fortuna2021well} and biased data distribution~\citep{swamy-etal-2019-studying,yin2021towards,fortuna-etal-2022-directions}. 
It prompts us to wonder \textit{what if} we apply the exact class definition to different datasets or annotate the same dataset using different class definitions. Unfortunately, manual hate speech annotation is time-consuming and very challenging. \citet{waseem-2016-racist} and \citet{caselli-etal-2020-feel} are among the few studies that re-annotated a  dataset, providing quantitative analysis or comparing the models' performance. However, such studies only apply to a single dataset. Moreover, the annotation is usually a one-shot effort, influenced by multiple factors related to the annotation task setup and knowledge of annotators. There is no way to attribute how much of the performance drop is due to incompatible class definitions and the data distribution \textit{separately}. 

In weakly-supervised models, we can interpret the category representation (and associated keywords) as the \textit{class definition}. Therefore, the class definition for the same taxonomy may differ depending on the dataset used to derive the category representation. Furthermore, we can approximate annotating a dataset with a different class definition by altering the category representation. 

We designed an ablation study to train X\nobreakdash-Class models using different combinations of datasets and class definitions. In Table~\ref{tab:cross-domain-deep-dive}, we present the results of three configurations in this study:\footnote{All experiments use the target taxonomy, and all documents are unlabeled.} 1) Using \textit{source}-dataset documents and category representations derived from the \textit{source} dataset (``X-Class$^{Agree}$ retrain'' in Table~\ref{tab:result-cross-domain}); 2) Using \textit{source}-dataset documents and category representations derived from the \textit{target} dataset; 3) Using \textit{target}-dataset documents and category representations derived from the \textit{target} dataset (``X-Class$^{Agree}$'' in Table~\ref{tab:result-in-domain}).

X-Class's cross-dataset performance substantially improved when provided with the category representation derived from the target dataset.\footnote{Its average recall in the ``Waseem $\rightarrow$ SBIC'' experiment decreased sharply mainly because the category representation for the ``Asian'' category is far from the document representation (the Waseem dataset does not contain documents related to ``Asian''). The model did not predict any document as ``Asian''.}
Only one factor is altered (either the category representation or the unlabeled training dataset) between the rows in Table~\ref{tab:cross-domain-deep-dive}. Therefore, we can conclude that the performance difference between rows \#1 and \#2 is due to different \textit{class definitions}, while the performance difference between rows \#2 and \#3 is due to different \textit{data distributions}.

\section{Conclusions and Future Work}
\label{sec:conclusions}

We applied extremely weakly-supervised methods to HS classification. We analyzed the transferability of HS classification models through comprehensive in-dataset and cross-dataset experiments and confirmed that weakly-supervised classification has several advantages over the traditional supervised classification paradigm. First, we can apply the algorithm across various HS datasets and domains with taxonomies that cannot be standardized using category mapping. Second, weakly-supervised models can readily utilize unlabeled documents in the target domain and do not suffer from domain mismatch problems. Lastly, weak supervision allows us to ``reannotate'' a labeled dataset using a different class definition to facilitate cross-dataset comparison, which was previously possible only at the cost of expensive manual annotation. 

The presented work is only the beginning of applying weak supervision to HS detection. We can utilize richer category representations than bag-of-keywords. However, such representations should be derived in an unsupervised or weakly-supervised manner to avoid depending on manually labeled datasets. A promising approach in this direction is \citet{shvets-etal-2021-targets}, which extracts HS targets and aspects relying on open-domain concept extraction. 

Lastly, we can study how well the model can generalize to previously unknown categories, a more challenging task often known as zero-shot classification~\citep{yin-etal-2019-benchmarking} or open-world classification~\citep{shu-etal-2017-doc}.

%\FloatBarrier
%\newpage

\section*{Limitations}

This study utilizes a monolingual pre-trained language model (PLM) in the English language (\texttt{bert-base-uncased}). Although the weakly-supervised classification methods are not limited to a particular language, we have not explored applying the method to another language. Social media language use may differ significantly from the data used to train the PLM. Moreover, the presence of code-switching~\citep{dogruoz-etal-2021-survey} may also degrade a monolingual PLM's performance. We explored a RoBERTa checkpoint continually trained with 60M English tweets~\citep{barbieri-etal-2020-tweeteval}.\footnote{\url{https://huggingface.co/cardiffnlp/twitter-roberta-base}} However, it does not yield better performance than BERT. We have not investigated whether it is due to the training regime or the dataset. 

Moreover, in this work, we focus on classifying hate speech (HS) categories/target groups instead of HS detection (detecting whether a post contains hate speech or not). To perform hate detection and classification, we can either combine our method with another HS detection model in a pipeline or use an adaptation of weakly-supervised text classification incorporating the ``Others'' category such as \citet{li-etal-2018-deep} or \citet{li2021misc}.

Due to limited space, we prioritized in-depth analysis instead of a comprehensive evaluation. Therefore, we selected only two datasets (and two-way cross-dataset classification). We are working in parallel on extending this work to a longer-form journal article to cover more datasets and experimental results.

Recent work on large language models (LLMs) demonstrated that when the parameters scale to a certain level, language models exhibit a drastically-increased performance in zero-shot classification~\citep{zhao2023survey}. We reported the performance of a moderately-sized \texttt{bert-large-uncased} zero-shot model because of limited computational resources and lack of access to commercial APIs. Larger language models will likely perform much better than this baseline.

Lastly, understanding HS sometimes requires cultural understanding or background knowledge. It may be difficult to determine the presence and category of HS when we take the post out of its context. For example, many ``Sexist'' posts in Waseem dataset are tweets related to the Australian TV show \textit{My Kitchen Rules} (MKR), and below is a tweet labeled as ``Sexist'':

\begin{quote}
Everyone else, despite our commentary, has fought hard too. It's not just you, Kat. \#mkr
\end{quote}

\section*{Ethics Considerations}

Although weak supervision requires only unlabeled documents, we demonstrated that the model might fail when the training dataset does not contain data related to a particular category or target group. It is especially concerning because the target groups are often minorities and under-represented. Therefore, we recommend against ``throwing'' a weakly-supervised algorithm on a dataset and hope the model will work. Instead, we should evaluate a model thoroughly before applying it to the real world, such as manually examining the model's predictions, behavioral testing the model using a checklist~\citep{ribeiro-etal-2020-beyond} or conducting unsupervised error estimation~\citep{jin-etal-2021-seed}.

\section*{Acknowledgements}
Leo Wanner and Alexander Shvets were partially supported by the European Commission under the grant number HE-101070278 and ISF-101080090. We thank Zeerak Talat for sharing the Waseem dataset with us and the anonymous reviewers for the careful reading and constructive feedback for us to improve the manuscript.

\bibliography{custom}
\bibliographystyle{acl_natbib}

\appendix

\section{Full List of Keywords in X-Class's Category Representation}
\label{appendix:full-keywords}

Table~\ref{tab:full-keywords-in-domain} shows the list of keywords in X-Class's category representation in the in-dataset setting (using the unlabeled documents and list of categories from the same dataset). Table~\ref{tab:full-keywords-out-domain} shows the list of keywords in X-Class's category representation in the cross-dataset setting (using the unlabeled Waseem dataset documents to induce category representations of SBIC dataset categories and vice versa).

\section{Reproducibility}
\label{appendix:reproducibility}

Table~\ref{tab:hyper-parameters} presents the hyper-parameters and their corresponding values to facilitate reproducing our result.

We use the \texttt{bert-large-uncased} model in HuggingFace as the base pre-trained language model for the zero-shot PET baseline. PET combines a \textit{pattern} (or prompt/instruction) with the input text and prompts the model to predict the mask token. Unlike open-ended prompting, PET uses a list of hand-crafted \textit{verbalizers} (candidate tokens). It classifies documents by assigning the category whose associated verbalizer receives the highest predicted probability. PET-style classification is especially beneficial for smaller PLMs, which do not possess a strong capability of instruction following~\citep{schick-schutze-2021-just,ouyang2022training}.

We hand-crafted patterns and verbalizers based on our understanding of the tasks (without fine-tuning). For Waseem dataset, we use the pattern ``<text> This hate speech is based on <mask>'' (verbalizers: gender/race), and for SBIC dataset ``<text> The target group of this hate speech is <mask>'' (verbalizers: women/black/Jews/gay/Muslims/Asian).

\section{Pseudo-Labeling}
\label{appendix:case-study}

Being able to accurately pseudo-label documents is crucial to the success of weak supervision. We report the accuracy of pseudo-labeling by various weakly-supervised methods in Table~\ref{tab:result-pseudo-label}.

\begin{table}[!htbp]
\centering 
\begin{tabular}{p{0.22\textwidth}cc}
\hline
\textbf{Dataset (Method)} & \textbf{Acc} & \textbf{P/R/F$_1$}\\ \cline{1-3}
Waseem & \textbf{99.1} & \textbf{98.4/99.2/98.9}  \\
- \textsc{W{\scriptsize E}STC{\scriptsize LASS}} & 77.8 & 77.9/80.1/77.4 \\
- LOTClass & 64.4 & 72.7/70.9/64.3 \\\cline{1-3}
SBIC & \textbf{93.0} & \textbf{89.1/92.8/91.1} \\
- \textsc{W{\scriptsize E}STC{\scriptsize LASS}} & 35.4 & 34.8/35.6/29.9  \\
- LOTClass & 51.8 & 32.4/26.3/24.5 \\
\cline{1-3}
SBIC $\rightarrow$ Waseem & 91.2 & 92.1/90.9/91.0 \\
\cline{1-3}
\end{tabular}
\caption{Pseudo-labeled dataset accuracy calculated against the gold-standard labels. The default method is X-Class unless otherwise specified. For the ``SBIC $\rightarrow$ Waseem'' setting, we use the category mapping in Table~\ref{tab:category-mapping} to derive the gold labels. We omit the ``Waseem $\rightarrow$ SBIC'' setting because we do not have gold labels.}
\label{tab:result-pseudo-label}
\end{table}

We can see that the accuracy of pseudo-labeled documents is consistent with the model's performance on the test dataset (Table~\ref{tab:result-in-domain}). Moreover, LOTClass and X-Class use the same underlying pre-trained language model (\texttt{bert-base-uncased}) in their final classifier, while \textsc{W{\scriptsize E}STC{\scriptsize LASS}} uses a more traditional convolutional neural networks architecture~\citep{kim-2014-convolutional}. The data pseudo-labeled by X-Class is substantially more accurate than the two baselines in both datasets. Comparing Table~\ref{tab:result-pseudo-label} and Table~\ref{tab:result-in-domain}, we can observe that the pseudo-labeling accuracy has a more significant impact on the final classifier's accuracy than the model architecture. 

We provide randomly sampled pseudo-labeled documents by X-Class in Table~\ref{tab:pseudo-sample-in-domain} (in-dataset) and Table~\ref{tab:pseudo-sample-out-domain} (cross-dataset). In general, the SBIC dataset contains more diverse and nuanced data. On the other hand, the Waseem dataset sometimes contains trivial slurs like ``... I'm not sexist ...''. The samples in the cross-dataset setting revealed that X\nobreakdash-Class tends to wrongly categorize original ``Sexist'' posts in the Waseem dataset (which mainly target women) as ``LGBT'' and ``Asian''.

\begin{table*}[!htbp]
\centering
\begin{tabularx}{\textwidth}{p{1.5cm}p{13.4cm}}
\cline{1-2}
\textbf{Class} & \textbf{Keywords} \\ \cline{1-2}
Sexist & sexist sexism misogynist sl*ts sl*t hypocrisy bigotry c*nts hypocrite bigoted pedophile filth c*nt phony barbarity scum bigot genocidal barbaric raping bitchy bigots rapist rapists blasphemy feminists mongering apostacy delusional trashy bimbos a*sholes skank retarded idiotic morons illiterate behead being sexual gays extremists sex islamophobia apostates whining self islamofascists beheads b*tches rape dudes beheading s*cking an enslave pure up common of a sassy vandaliser gender by feminist \\
\rowcolor{Gray}
Racist & racist racists racism naziphobia fascist oppression hateful hatred semitic imperialist hating race imperialism genocide inhuman vile ideology violent murderous violence anti nazism vileness brutal propaganda nazis terrorist filthy disgusting radical murdering terrorists hate abuse attacking islamists islamolunatic islamolunatics minority murderers domination jihad terrorism islamist westerners evil killing attack against hated atheists political terror murder culture minorities religious lunatics human conspiracy population hatewatch killings secular religion force cult \\ \cline{1-2}
Women & women woman female females girls ladies ch*cks wives men feminist lady girl chick feminists feminine males male gender feminism whores blonde virgins bitches guys hookers prostitutes sl*ts mens wh*re sl*t b*tch p*ssy prostitute virgin couples d*cks breast moms c*nts girlfriend wife sisters dudes attractive sexy betas partners she her beautiful genders lovers normies mothers boys man chads adult couple them fathers mensrights normie assholes they body someone bodies looking v*ginas loser dyk*y sister ones femaloid self mate material raped hooker \\
\rowcolor{Gray}
Black & black white colored blacks whites n*gro african negroes negros racial race racist races minorities color africans n*groids minority n*groid racism mixed brown n*ggers skinned blackman slaves peoples ghetto discrimination n*gger people whitey africa red yellow dark savages individuals civil poor disabled blind gorillas savage human folk nonwhite left lynching slavery diversity worthless folks south gorilla majority violent dirty green cotton slave \\ 
Jewish & jewish jews jew synagogue rabbi israel zionist semitic holocaust kosher auschwitz nazi goyim german aryan germans nazis ethiopian germany hitler concentration ash \\ 
\rowcolor{Gray}
LGBT & gay homosexual gays homosexuals homosexuality lesbian lesbians queer transgender homophobic sexuality sexual h*mo queers transgenders masculine sexism sexist trans sex sexually straight dating anal dyke dykes penis marriage rape erection pubic openly pedophile porn nude hiv aids raping interracial relationships relationship genitals boyfriends pedophilia objectifying bi std naked d*ck cocks date misogynist misogyny threesome masturbating shaming stoned v*gina assault bestiality c*nt f*cks rapist genital hot c*ck \\
Muslim & muslim muslims islamic islam mosque mosques arabic quran arab muhammad mohammed shia prophet religion terrorists christian religious allah saudi christians terrorist pakistani arabia ali terrorism pakistan prophets bombers isis syria al qaeda banislam radical camels mass bomber bombing church refugees iran suicide iraq middle faith mosul abdul converted jesus akbar military bomb nations militant pray god kkk militia attacks bible propaganda attack \\ 
\rowcolor{Gray}
Asian & asian asians chinese oriental korean japanese american vietnamese indian ethnic mexican americans english latina china eastern foreign exotic european koreans pacific russian north indians spanish russians thai east korea japan country america french cultural western irish countries cuban international nigerian chinaman culture british primitive aged ape inner refugee alien older states europe united animal fat nationality usa russia armed old ignorant special city iq traitor eating animals hungarian food intelligent modern state vietnam rice  \\\cline{1-2}
\end{tabularx}
\caption{Full list of keywords in X-Class's category representation mined from \textit{in-dataset} setting.}
\label{tab:full-keywords-in-domain}
\end{table*}

\begin{table*}[!htbp]
\centering
\begin{tabularx}{\textwidth}{p{1.5cm}p{13.4cm}}
\cline{1-2}
\textbf{Class} & \textbf{Keywords} \\ \cline{1-2}
Sexist & sexist sexism homophobic misogyny misogynist hypocrisy sl*ts sl*t c*nts sl*tty degenerate pedophile pedophilia lesbians sexual masculinity bestiality stereotypical shaming whores feminists masturbating mutilation trashy objectifying homosexuals sexually patriarchy misandry raping c*nt rapist hypocritical gays discriminated genital degeneracy unoriginal a*sholes retarded queers virgins disgusting cannibalism self kinky barbarity promiscuity genitals f*cks rape \\
\rowcolor{Gray}
Racist & racist racism racial discrimination race ethnic races blacks black colored white whites n*gro african negroes minority asians minorities oppression diversity negros ghetto n*groid n*groids mixed cultural peoples africans n*ggers color semitic americans american culture asian individuals people savages savage violence slavery n*gger mixing transgenders mass skinned worthless queer slaves \\ \cline{1-2}
Women & women woman female females ladies girls feminine feminist feminists feminism womens gender male men girl lady ch*cks blonde blondes males femininity mens wives guys ch*ck wife yesallwomen b*tches daughters her she stars b*tchy girlfriend body b*tch sister feminismisawful announcers promogirls sportscasters bodies models they themselves refs ones them couples someone diva their sjw mother \\
\rowcolor{Gray}
Black & black white blacks whites racists racist race minorities minority racism africans oppressed americans oppression people population human \\ 
Jewish & jewish jews jew judaism israel palestinian zionist palestinians israelis israeli palestine semitic semitism hamas gaza holocaust nazis nazi egyptians\\ 
\rowcolor{Gray}
LGBT & gay gays sexual sex sexism sexists sexist rape raping misogynist rapists reproductive misogyny pedophile rapist genitals sl*ts sl*t raped c*nts assault dudes masculinity porn boys shaming c*nt hypocrisy v*gina bigotry rapes bigoted hypocrites hateful haters stereotype openly bimbos wh*re abuse misandrist \\
Muslim & muslim muslims islamic islam islamist sunni religious islamists jihadi jihadis arab arabs mosques shia quran jihad religion muhammad mohammed taqiyya allah terrorist terrorists prophet believers religions christian hadiths sharia baghdadi secular caliphate hadith saudis saudi pakistani imam christians terrorism islamolunatics isis islamofascists arabian arabia umar extremists hindus pakistan taquiyya medina qurans mullah sunnah westerners  \\ 
\rowcolor{Gray}
Asian & asian intelligent attractive ignorant young pretty dumb hot rich fat ugly stupid smart tough looking crazy insane blond selfish common brainwashed correct biased clever annoying childish being most hating seeing old beautiful terrible killer self innocent a everydaysexism friendly average ridiculous idiotic extremely poor good bad flawed decent great low simple nice an legit out safe trash doing useless awful corrupt funny sick strong other known working many making best no  \\\cline{1-2}
\end{tabularx}
\caption{Full list of keywords in X-Class's category representation mined from \textit{cross-dataset} setting.}
\label{tab:full-keywords-out-domain}
\end{table*}

\begin{table*}[!htbp]
\centering
\begin{tabularx}{\textwidth}{p{2.9cm}p{2.8cm}p{8.9cm}}
\cline{1-3}
\textbf{Hyper-parameter} & \textbf{Value} & \textbf{Description} \\ \cline{1-3}
\texttt{random\_seed} & 42 & The fixed random seed. Used to split the dataset and initialize parameters. \\
\rowcolor{Gray}
\texttt{lm\_ckp} & bert-base-uncased & The pre-trained language model checkpoint used to derive document representations.\\
\texttt{clf\_ckp} & bert-base-uncased & The pre-trained language model checkpoint used to fine-tune the final classifier. Used in both supervised and weakly-supervised settings.\\
\rowcolor{Gray}
\texttt{min\_freq} & 5 & Minimum frequency of a word to be included in the vocabulary. \\
\texttt{T} & 100 & Maximum terms to include in the category representation. \\
\rowcolor{Gray}
\texttt{cluster\_method} & gmm & Method to perform document class alignment in X-Class. We use the default Gaussian Mixture Model with tied covariance. \\
\texttt{pca\_dim} & 64 & Dimension of principal component analysis before performing clustering. \\
\rowcolor{Gray}
\texttt{conf\_threshold} & 0.5 & The percentage of most confident documents assigned by GMM to include in the pseudo-labeled training set. \\\cline{1-3}
\texttt{max\_len}$^{*}$ & 64 & The maximum number of tokens of the input posts. Input longer than it will be truncated. \\
\rowcolor{Gray}
\texttt{batch\_size}$^{\ddag}$ & 64 & The training batch size. \\
\texttt{n\_epochs}$^{\ddag}$ & 6 & The number of training epochs. \\
\rowcolor{Gray}
\texttt{learning\_rate}$^{\ddag}$ & 2e-5 & \\
\texttt{weight\_decay}$^{\ddag}$ & 0.05 & \\
\rowcolor{Gray}
\texttt{lr\_schedule}$^{\dag}$ & cosine w/ warmup & Learning rate schedule. Cosine learning rate schedule with 1/3 steps warm-up.\\\cline{1-3}
\end{tabularx}
\caption{Full list of hyper-parameters. The first block is hyper-parameters related to X-Class. The second block is hyper-parameters related to the classifier fine-tuning. * denotes values set based on data analysis. \ddag\ denotes values set by manual hyperparameter tuning. \dag\ denotes values chosen based on our experience but are not tuned. All the rest of the parameters default from the X-Class repo.}
\label{tab:hyper-parameters}
\end{table*}

\begin{table*}[!htbp]
\centering
\begin{tabularx}{\textwidth}{p{1.2cm}p{11.8cm}c}
\cline{1-3}
\textbf{Class} & \textbf{Posts} & \textbf{Correct?}  \\ \cline{1-3}
Sexist & AbuserDynamics Girls should count themselves lucky if they got a guy that still shouts when his mad, means he cares & $\checkmark$ \\
& I have a stupid man crush on Lloyd and I can't help it. mkr2015 mkr sexylloydmkr & $\checkmark$ \\
& RT This lady is a terrible official. Not because she's a lady though. I'm not sexist. But because she's a stupid b*tch. Cavs & $\checkmark$ \\
\rowcolor{Gray}
Racist & RT CNN reports on the murders of homosexuals by the Islamic State, but the problem isn t just ISIS, it s Islam & $\checkmark$ \\ 
\rowcolor{Gray}
& Yes, there is even more rape in Muslim countries but it is not reported. & $\checkmark$ \\
\rowcolor{Gray}
& For example, the Quran says leaving Islam is not accepted. The Hadiths say kill anyone who leaves. & $\checkmark$ \\
\cline{1-3}
Women & Molly make b*tches crazy & $\checkmark$ \\
& The emoji I use when I'm flirting wit b*tches gt & $\checkmark$ \\
& Rule number two don't trust nobody especially a b*tch with a hookers body & $\checkmark$ \\
\rowcolor{Gray}
Black & It's funny when u get a black female mad they start acting ghetto then a b*tch & $\checkmark$ \\
\rowcolor{Gray}
& White bitches be lovin da colored men & \ding{55} \\
\rowcolor{Gray}
& Those ratchet broads don't represent me but YOUR white a*s can't talk trash about them 'cause you on racist sh*t ``Black''' & $\checkmark$ \\
Jewish & What was the Jewish pirate's nickname? Hooknose. & $\checkmark$  \\
& Do you know why do Jewish kids wait behind the chimney? Because they're waiting for their parents . & $\checkmark$  \\
& Why do Jews hate a*sholes? The can't stand gas chambers. & $\checkmark$  \\
\rowcolor{Gray}
LGBT & Any man even considering plastic surgery is gay gay gay imagine ya boy say ``I'd get more hoes then u after I get these new lips'' & $\checkmark$ \\
\rowcolor{Gray}
& do it i f*cking dare you f*ggot & $\checkmark$ \\
\rowcolor{Gray}
& What do you call a gay drive by? A fruit roll up. & $\checkmark$ \\
Muslim & RT Fuck Islam you pigfucking trash troll. & $\checkmark$ \\
& Bomb your house, Then f*ck your b*tch. All hail Al Qaeda & $\checkmark$ \\
& What did the Muslim say after eating a bacon sandwich? That's the way Allah Allah I like it. & $\checkmark$ \\
\rowcolor{Gray}
Asian & My economics professor just called these asians retards & $\checkmark$ \\
\rowcolor{Gray}
& Korean idols are perfectly handsome, cute, funny, stylish, hot, know how to dance, have a wonderful voices. & \ding{55}\\ 
\rowcolor{Gray}
& The Stock Exchange I like the NYSE just like the Ethiopian population count. Going down faster than ever. & \ding{55}\\ 
\cline{1-3}
\end{tabularx}
\caption{Randomly sampled pseudo-labeled examples for each category in the in-dataset setting.}
\label{tab:pseudo-sample-in-domain}
\end{table*} 

\begin{table*}[!htbp]
\centering
\begin{tabularx}{\textwidth}{p{1.2cm}p{11.8cm}c}
\cline{1-3}
\textbf{Class} & \textbf{Posts} & \textbf{Correct?}  \\ \cline{1-3}
Sexist & on sale a*s hoes & $\checkmark$ \\
& Molly make b*tches crazy & $\checkmark$ \\
& This n*ggga said I be branding b*tches & \ding{55} \\
\rowcolor{Gray}
Racist & RT Wow the stupid n*gger in LeBron really came out there & $\checkmark$ \\ 
\rowcolor{Gray}
& My Moor friends,no not black friends,but Moor friends said N*gger came from Nigeria...You are so lost..Stop tagging me... & $\checkmark$ \\
\rowcolor{Gray}
& RT Remember the ``yellow badge'' Nazis used? Israel is making Muslim women carry a yellow badge order to pray in Al Aqsa. h & $\checkmark$ \\
\cline{1-3}
Women &  RT I'm no sexist but the last thing I wanna read about is women's, football or cricket on the sky sports news app! controve & $\checkmark$ \\
& RT Then I guess Feminism is just a sideshow as much as WWE wrestling in general.. Irony is off the c & $\checkmark$ \\
& Are you even a real person? I'm not sexist. But Men are superior to women  & $\checkmark$ \\
\rowcolor{Gray}
Black & Can't forget it...never heard about it... & \ding{55} \\
\rowcolor{Gray}
& ...with a flat face. The nose a bay window. & \ding{55} \\
\rowcolor{Gray}
& But look at the reality disconnect. Burak says he is for freedom and against all slavery while at the ... & $\checkmark$ \\
Jewish & Max Blumenthal is bad mouthing you. Not enough room at the top for all the self genocidal Jews. Israel Palestine & $\checkmark$  \\
& The job Mohammed set Muslims is not done while Israel exists. & $\checkmark$  \\
& The Jews of Europe should just come to the US. Then the Europeans can allow Islam to take them backwards. & $\checkmark$  \\
\rowcolor{Gray}
LGBT & RT I'm not sexist but right now I hate girls !!!! & \ding{55} \\
\rowcolor{Gray}
& RT This is not sexist but I want to punch both of the girls from broad city workaholics & \ding{55}\\
\rowcolor{Gray}
&  RT This is why girls don't play football. Someone's feelings get hurt and boom, it's out of hand. Go ahead and call me sexist, & \ding{55} \\
Muslim & You didn't recognize the irony of me using your method because you are an ignorant Muslim. & $\checkmark$ \\
& And you lie again. The majority of Muslims were forced into it. & $\checkmark$ \\
& RT Arab slave trade 140 to 200 million non Muslim slaves from all colors and nationalities still happening today!  & $\checkmark$ \\
\rowcolor{Gray}
Asian &  Someone really needs to get the sniffer dogs onto Kat offherlips MKR  & \ding{55}\\
\rowcolor{Gray}
& MKR anyone can cook from a can girls. & \ding{55}\\ 
\rowcolor{Gray}
& Kat you don't look suspicious at all! MKR& \ding{55}\\ 
\cline{1-3}
\end{tabularx}
\caption{Randomly sampled pseudo-labeled examples for each category in the cross-dataset setting.}
\label{tab:pseudo-sample-out-domain}
\end{table*} 

\end{document}